\documentclass[runningheads]{llncs}
\usepackage{algorithm}
\usepackage{algorithmic}

\usepackage{newfloat}
\usepackage{listings}
\floatstyle{ruled}
\newfloat{listing}{tb}{lst}{}
\floatname{listing}{Listing}

\usepackage{tikz}
\usepackage{tkz-berge}

\usepackage{hyperref}
 \definecolor{lime}{HTML}{A6CE39}
 \DeclareRobustCommand{\orcidicon}{
 	\begin{tikzpicture}
 	\draw[lime, fill=lime] (0,0)
 	circle [radius=0.16]
 	node[white] {{\fontfamily{qag}\selectfont \tiny ID}};
 	\draw[white, fill=white] (-0.0625,0.095)
 	circle [radius=0.007];
 	\end{tikzpicture}
 }
 \renewcommand{\orcidID}[1]{\href{https://orcid.org/#1}{\ensuremath{\orcidicon}}}

\usepackage[utf8]{inputenc}
\usepackage{float} 
\usepackage[left=3cm,right=3cm,top=3cm,bottom=3cm]{geometry}
\usepackage{graphicx}
\begin{document}
\title{Generalized Nested Rollout Policy Adaptation with Dynamic Bias for Vehicle Routing}
\author{
    Julien Sentuc\inst{1}\orcidID{0000-0003-4504-6128} \and
    Tristan Cazenave\inst{1}\orcidID{0000-0003-4669-9374} \and
    Jean-Yves Lucas\inst{2}\orcidID{0000-0002-2482-5686}
}
\authorrunning{J. Sentuc et al.}
\institute{LAMSADE, Université Paris-Dauphine, PSL, CNRS\\
\email{Julien.Sentuc@dauphine.eu, Tristan.Cazenave@dauphine.psl.eu} \and
OSIRIS department, EDF Lab Paris-Saclay, Electricité de France, France\\
\email{Jean-Yves.Lucas@edf.fr}}
\maketitle              
\begin{abstract}
In this paper we present an extension of the Nested Rollout Policy Adaptation algorithm (NRPA), namely the Generalized Nested Rollout Policy Adaptation (GNRPA), as well as its use for solving some instances of the Vehicle Routing Problem. We detail some results obtained on the Solomon instances set which is a conventional benchmark for the Capacitated Vehicle Routing Problem with Time Windows (CVRPTW). We show that on all instances, GNRPA performs better than NRPA. On some instances, it performs better than the Google OR Tool module dedicated to VRP.
\end{abstract}

\section{Introduction}
Monte Carlo Tree Search \cite{Kocsis2006,Coulom2006} has been successfully applied to many games and problems \cite{BrownePWLCRTPSC2012}. It originates from the computer game of Go \cite{Bouzy01} with a method based on simulated annealing. 

Nested Monte Carlo Search \cite{CazenaveIJCAI09} is a recursive algorithm which uses several search levels, memorizing the best sequence at each level. At each stage of the search, the move with the highest score at the lower level is played by the current level. At each step, a lower-level search is launched for all possible moves and the move with the best score is memorized. At level 0, a Monte Carlo simulation is performed, random decisions are made until a terminal state. At the end, the score for the position is returned. NMCS has given good results on many problems like puzzle solving, single player games \cite{Mehat2010}, cooperative path finding or the inverse folding problem \cite{portela2018unexpectedly}.

Based on the latter, the Nested Rollout Policy Adaptation (NRPA) algorithm was introduced \cite{Rosin2011}. It also uses multiple levels of search, memorizing the best sequence of moves. It consists of the online learning of a playout policy using the best sequence in a nested search. NRPA achieved world records in Morpion Solitaire and crossword puzzles. It has been applied to many problems such as object wrapping \cite{edelkamp2014monte}, traveling salesman with time window \cite{cazenave2012tsptw,edelkamp2013algorithm}, vehicle routing problems \cite{edelkamp2016monte,Cazenave2020VRP} or network traffic engineering \cite{DangMonteCarlo2021}.

This work aims at adapting the NRPA algorithm and its extension GNRPA (Generalized Nested Rollout Policy Adaptation) \cite{Cazenave2020GNRPA} to the Vehicle Routing Problem (VRP). We demonstrate that GNRPA with our bias outperforms NRPA.

This paper is organized as follows. Section 2 presents the VRP and the 56 Solomon instances that we used as a benchmark.  Section 3 describes the NRPA and  GNRPA algorithms. Section 4 presents the experimental results. Finally, the last section concludes. 

\section{The Vehicle Routing Problem}
The Vehicle Routing Problem is one of the most studied optimization problems. It was first introduced in 1959 in "The Truck Dispatching Problem" \cite{dantzig_ramser59} by G.B. Dantzig and J.H. Ramser. This problem consists of finding a set of optimal paths given a certain number of delivery vehicles, a list of customers or places of intervention as well as a set of constraints. This problem is therefore an extension of the traveling salesman problem. In its simplest version, all vehicles leave from the same depot. The goal is then to minimize the objective function, generally defined by 3 criteria given in order of importance: the number of customers that were not serviced, the number of vehicles used, and the total distance traveled by the whole set of vehicles. These 3 criteria may be assigned specific weights in the objective function, or a lexicographic order can be taken into account. The vehicle routing problem is NP-hard, so there is no known algorithm able to solve any instance of this problem in polynomial time. Although exact methods such as Branch and Price exist, approximate methods are nonetheless useful for solving difficult instances.

\subsection{VRP variations}

Many companies find themselves faced with the vehicle routing problem \cite{Cazenave2021Policy}. Many variations of the vehicle routing problem have therefore been created.%

Capacitated Vehicle Routing Problem (CVRP) adds a demand to each customer (e.g., the number of parcels they have purchased) and a limited carrying capacity for all vehicles. It is impossible for a vehicle to restock.

The Vehicle Routing Problem with Time Windows (VRPTW) implies to serve
each customer within a given time window (possibly different for each customer).The depot also has a time window, thus limiting the duration of the tour. 

CVRPTW is a combination of CVRP and VRPTW. Figure 1 shows an illustration of the CVRPTW problem. In this example, each car initially has a capacity of 100; each vertex has a time window in which a vehicle must arrive in place. In addition, each customer is assigned two values. First, a demand, denoted here by q, corresponding to the quantity that the vehicle must supply. The second corresponds to the service duration, denoted here by d.

\begin{figure}[H]
    \centering
    \includegraphics[width=0.5\textwidth,keepaspectratio]{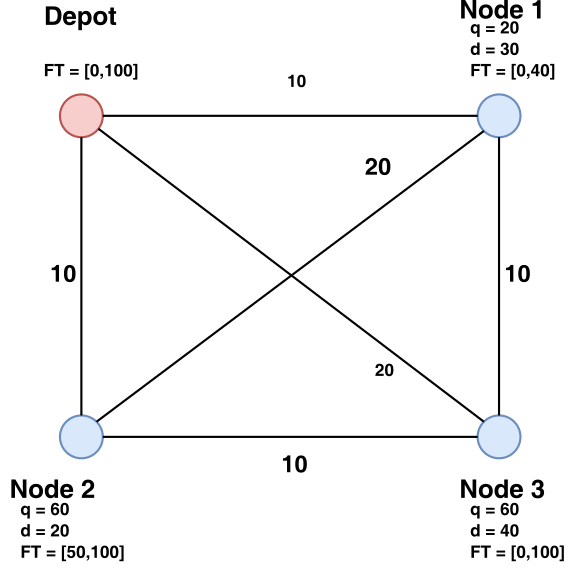}
    \caption{Graph Example for CVRPTW}
    \label{fig:my_label}
\end{figure}

\subsection{Solomon Instances}
To evaluate the different algorithms, this work uses the 1987 Solomon instances \cite{Solomon} for the CVRPTW problem. Solomon's instances are the main benchmark for CVRPTW. The benchmark is composed of 56 instances, all consisting of a depot and 100 customers with coordinates included in the interval [0,100]. Vehicles start their tours with the same capacity defined in the instance. A time window is defined for each client as well as for the depot. The distances and the durations correspond to the Euclidean distances between the geometric points. 

Solomon's problems are divided into six classes, each having between 8 and 12 instances. For classes R1 and R2, the coordinates are generated randomly, while for classes C1 and C2 are cluster based. A mixture of random and cluster structures is used for the problems of classes RC1 and RC2. The R1, C1, and RC1 problem sets have a short time horizon and only allow a few clients per tour (typically up to 10). On the other hand, the sets R2, C2 and RC2 have a long time horizon, allowing many customers (more than 30) to be serviced by the same vehicle.

\subsection{Some methods commonly used for solving a VRP}

A wide range of methods have been used to solve the VRP. We limit ourselves to briefly describe some of them.

\subsubsection{Column Generation}
 is a particular case of Dantzig-Wolfe decomposition. It is used in the framework of Mixed Integer Linear Programming, in order to find optimal solutions to difficult optimization problems. It is based on the fact that, usually, an optimal solution of a linear program contains many non-basic variables, meaning that their value is zero in the solution. Thus, in theory, only a subset of the variables needs to be taken into account. However, the appropriate subset of variables has to be found. To achieve this goal, the method breaks down the problem into two parts, the $master problem$ (the original problem with a reduced set of variables), and the $subproblem$  which identifies a variable with a negative reduced cost. This variable is then added to the subset of variables taken into account by the $master problem$, and the process is repeated until no new variable can be added to the subset. For a VRP, the $subproblem$ is usually modeled as the shortest path problem with resources constraints.  

\subsubsection{Heuristic methods}
does not aim at always finding an optimal solution. It rather aims at finding quickly (very) good solutions. There exists two main types of heuristics, construction heuristics that build solutions through iterations, and descending heuristics that search for a local optimum from a given solution. The cluster-first, route-second heuristic \cite{Fisher81} and the savings heuristic \cite{Clarke64} are two exemples of heuristics used for solving vrp problems.
\subsubsection{Metaheuristic methods}
are higher level procedures used to escape local optimums and find better solutions. This is achieved by moving around in the search area and exploring promising regions. The most commonly used metaheuristics for solving VRP problems are the particle swarm optimization \cite{Marinakis2013,Alinezhad}, simulated annealing \cite{Chiang96}, genetic algorithm \cite{IEECAIA1991,Prasetyo}, Tabu search \cite{Cordeau97,Xia}, ant colony algorithm \cite{Zheng_2020}. Another commonly used metaheuristic is the local search: one starts from an initial solution to the problem, possibly not fulfilling all constraints. Then at each step, a neighborhood of the current solution is built. This neighborhood contains solutions obtained by slightly modifying the current solution, that is, by changing some variables assignments. The best neighbor solution (with regard to the objective function or to the number of violated constraints) is kept and becomes the new current solution. This process is repeated until a time limit is reached, or a satisfactory solution is found.

\section{Monte Carlo Search}
This section presents the NRPA algorithm as well as its generalization GNRPA. The different variations studied in this work are then presented. Finally, this section describes the way CVRPTW problems are modeled.
\subsection{NRPA}
The Nested Rollout Policy Adaptation (NRPA) \cite{Rosin2011} algorithm is an effective combination of nested Monte Carlo search \cite{cazenave09bus} and policy learning. NRPA holds world records for Join Five and crosswords puzzles. The algorithm performs Gibbs sampling, choosing actions with a probability given by the softmax function. \\
Each action is encoded as an integer, which is the index of its weight in the policy table. Algorithm 1 therefore sequentially constructs a random solution biased by the weight of the moves until it reaches a terminal state. The policy is then adapted to the best solution found at the current level. The algorithm starts by saving the current policy into a temporary policy array named polp before modifying it. The policy copied into polp will be modified by the Adapt function, while the current policy will be used to calculate the probabilities of possible hits. After modification of the policy, the current policy is replaced by polp. The principle of adaptation is to increase the weight of the chosen moves and to decrease the weight of the other possible moves by an amount proportional to their probabilities of being played. In NRPA, each level takes a policy as input and returns a sequence and its associated score. At any level $>$ 0, the algorithm makes numerous recursive calls to lower levels, adapting the policy each time with the best solution to date. It should be noted that the changes made to the policy do not affect higher levels. At level 0, we return the sequence obtained by algorithm 1 as well as its associated score. The NRPA algorithm therefore strikes a balance between exploration and exploitation.  It exploits by shifting his weights to the best current solution and explores using Gibbs sampling at the lower level.
This is a general algorithm that has been shown to be effective for many optimization problems.\\
The idea of adapting a simulation policy has been applied successfully for many games such as Go \cite{graf2015}.
It should be noted that in the case of the VRP, we aim at minimizing the score (related to a set of penalties). $bestScore$ is therefore initialized to $+\infty$ and we update it each time we find a new $result$ such that $result \leq bestScore$. \\
Let us also mention NRPA with initialisation, that is an extension of NRPA. In this varition, the original weights in the policy array are not uniformely set to 0, but to an appropriate value according to a heuristic relevant to the problem to solve. 

\subsection{GNRPA}
Let $w_{ic}$ be the weight associated with move $c$ in step $i$ of the sequence. In NRPA, the probability of choosing move $ c $ at the index $ i $ is defined by: 
$$ p_{ic} = \frac{e^{w_{ic}}}{\sum_k{e^{w_{ik}}}} $$

GNRPA \cite{Cazenave2020GNRPA} generalizes the way the probability is calculated using a temperature $ \ tau $ and a bias $ \ beta_ {ic} $. The probability of choosing the move $ c $ at the index $ i $ then becomes: 
$$ p_{ic} = \frac{e^{\frac{w_{ic}}{\tau}+\beta_{ic}}}{\sum_k{e^{\frac{w_{ik}}{\tau}+\beta_{ik}}}} $$
By taking $ \tau = 1 $ and $ \beta_ {ik} = 0 $, we find the formula for NRPA. However, it is sometimes more practical to use $ \beta_ {ij} $ biases than to initialize the weights, as will be mentioned below.

\begin{algorithm}
\begin{algorithmic}[1]
\STATE{playout ($state$, $policy$)}
\begin{ALC@g}
\STATE{$sequence$ $\leftarrow$ []}
\WHILE{true}
\IF{$state$ is terminal}
\RETURN{(score ($state$), $sequence$)}
\ENDIF
\STATE{$z$ $\leftarrow$ 0}
\FOR{$m \in$ possible moves for $state$}
\STATE{$o [m] \leftarrow e^{\frac{policy[code(m)]}{\tau} + \beta(m)}$}
\STATE{$z \leftarrow z + o [m]$}
\ENDFOR
\STATE{choose a $move$ with probability $\frac{o [move]}{z}$}
\STATE{$state \leftarrow play (state, move)$}
\STATE{$sequence \leftarrow sequence + move$}
\ENDWHILE
\end{ALC@g}
\end{algorithmic}
\caption{\label{PLAYOUTI}The generalized playout algorithm}
\end{algorithm}

\begin{algorithm}
\begin{algorithmic}[1]
\STATE{Adapt ($policy$, $sequence$)}
\begin{ALC@g}
\STATE{$polp \leftarrow policy$}
\STATE{$state \leftarrow root$}
\FOR{$b \in sequence$}
\STATE{$z \leftarrow 0$}
\FOR{$m \in$ possible moves for $state$}
\STATE{$o [m] \leftarrow e^{\frac{policy[code(m)]}{\tau} + \beta(m)}$}
\STATE{$z \leftarrow z + o [m]$}
\ENDFOR
\FOR{$m \in$ possible moves for $state$}
\STATE{$polp [code(m)] \leftarrow polp [code(m)] - \frac{\alpha}{\tau}(\frac{o[m]}{z} - \delta_{bm})$}
\ENDFOR
\STATE{$state \leftarrow play (state, b)$}
\ENDFOR
\STATE{$policy \leftarrow polp$}
\end{ALC@g}
\end{algorithmic}
\caption{\label{ADAPTI}The generalized adapt algorithm}
\end{algorithm}

\begin{algorithm}
\begin{algorithmic}[1]
\STATE{GNRPA ($level$, $policy$)}
\begin{ALC@g}
\IF{level == 0}
\RETURN{playout (root, $policy$)}
\ELSE
\STATE{$bestScore$ $\leftarrow$ $-\infty$}
\FOR{N iterations}
\STATE{(result,new) $\leftarrow$ NRPA($level-1$, $policy$)}
\IF{result $\geq$ bestScore}
\STATE{bestScore $\leftarrow$ result}
\STATE{seq $\leftarrow$ new}
\ENDIF
\STATE{policy $\leftarrow$ Adapt (policy, seq)}
\ENDFOR
\RETURN{(bestScore, seq)}
\ENDIF
\end{ALC@g}
\end{algorithmic}
\caption{\label{GNRPA}The GNRPA algorithm.}
\end{algorithm}

\subsection{Optimizations of GNRPA}
\subsubsection{Avoid the copy of the policy}
In the Adapt algorithm for NRPA and GNRPA, the sum of the gradients is calculated for the entire sequence and then applied. The policy is first copied into a temporary policy array, which is modified using the gradient and the original policy before being copied back into the policy. When the number of possible codes is large, copying the policy can be costly. We can avoid this by first making a loop to compute the probabilities of each move of the best sequence, and then apply the gradient directly to the policy for each state number and each code.
\subsubsection{Avoid calculating again the possible moves}
For some problems, the computation of the possible moves can be costly. In order to avoid recalculating the possible moves for the best playout in the Adapt function, we can memorize them in the playout function, where they have already been calculated. The optimized playout algorithm memorizes in a matrix $code$ the codes of the possible moves during a playout. The cell $code[i][m]$
contains the code of the possible move of index $m$ at state number $i$ of the best sequence. The $index$ array stores the index of the best move code for each step of the best sequence.
\subsubsection{Avoid calculating again the biases}
In some cases, the computation of the bias for all possible moves can be costly. As for the possible moves, we store the values of the bias in a $\beta$ matrix. The cell $\beta[i][m]$
contains the bias of the possible move of index $m$ at state number $i$.

\begin{algorithm}
\begin{algorithmic}[1]
\STATE{Adapt ($policy, code, index$)}
\begin{ALC@g}
\FOR{$i \in [0,len(index)[$}
\STATE{$z [i] \leftarrow 0$}
\FOR{$m \in [0,len(code [i])[$}
\STATE{$o [i] [m] \leftarrow e^{\frac{policy[code[i] [m]]}{\tau} + \beta[i][m]}$}
\STATE{$z [i] \leftarrow z [i] + o [i] [m]$}
\ENDFOR
\ENDFOR
\FOR{$i \in [0,len(index)[$}
\STATE{$b \leftarrow index [i]$}
\FOR{$m \in [0,len(code [i])[$}
\STATE{$policy [code [i] [m]] \leftarrow policy [code [i] [m]] - \frac{\alpha}{\tau}(\frac{o [i] [m]}{z |i]} - \delta_{bm})$}
\ENDFOR
\ENDFOR
\end{ALC@g}
\end{algorithmic}
\caption{\label{OGADAPT}The optimized generalized adapt algorithm}
\end{algorithm}

\subsection{Modeling the problem}
There are some design choices that have to be made when implementing GNRPA for CVRPTW. 

The first choice is to decide how to encode the possible moves in the policy. The first option consists of encoding a move as a place of departure, a place of arrival and a vehicle by the formula $ code (i_d, i_a, v) = i_d + n * i_a + n ^ 2 * v $ where $ i_d $ is the index of the customer/deposit at departure, $ i_a $ the index of arrival, $v$ the index of the vehicle and $n$ the number of customers + 1 (the deposit). The second option consists of taking only the departure and the arrival nodes to encode the move. Although the first possibility prevents the forced moves at the end of the sequence from deteriorating the policy, the knowledge learned is spread among the different vehicles. Logically, we obtained significantly better results with option 2. It should be noted that option 1 is relevant when the  different vehicles do not have the same characteristics (or not the same start, or not the same cargo, etc.). 

Another design choice is to define a score function to evaluate the simulations. A first approach \cite{Cazenave2021Policy} consists of using 3 weighted penalties: the number of unserviced customers (associated to high weighting), the number of vehicles used (associated to average weight) and the total distance traveled (associated to low weight). Moves that do not respect the time window constraints are not considered. The score function is defined by the following formula:
$$Unvisited * 10 ^ 6 + nVehicles * 10 ^ 3 + distance $$
The second approach consists of accepting the moves which violate constraints while penalizing them, as in \cite {cazenave2012tsptw}. The score function then becomes:
$$Unvisited * 10 ^ 6 + nVehicles * 10 ^ 3  + distance$$
$$+ \Omega (P) * 10 ^ 6 $$
where $ \Omega (P) $ is the number of violated constraints. In this paper, we implemented the first approach.

Finally, a final design choice consists of allowing, or not allowing, the vehicle to return to depot whereas it still could service more customers in the same tour. Allowing the vehicle to return prematurely to the depot might help in some cases to escape from local optima. However, our experiments showed that it generally produced bad solutions, so we did not retain this possibility. 

\subsection{Use of the bias}
As mentioned earlier, the bias is dynamic. The current state can thus be taken into account when calculating the bias. The previous works which studied the application of NRPA to the problem of vehicle routing with time window used an initialization of the policy weights from the distances between two places \cite{EdelNRPA}. Here we used a dynamic bias inspired from the Solomon constructive heuristic \cite{Solomon}. It's made up of 3 parts. First, the distance, like previous works. Second, the waiting time on arrival. Third, the "lateness". This consists of penalizing an arrival too early in a time window. The formula used for the bias is thus :
$$\beta_{ic}=  w_1 * \beta_{distance} + w_2 * \beta_{waiting} + w_3 * \beta_{lateness},$$
with $w_1,w_2,w_3>0$.

\cite{cazenave2012tsptw} proposed a similar heuristic for the TSPTW.

Various formulas have been proposed for weight initialization. For example, Tristan Cazenave used the formula $c\times \frac{d_{ij}-min}{max-min}$ in \cite{Cazenave2020GNRPA} for the Traveling Salesman with Time Windows problem (TSPTW) and $c \times \frac{1}{d_{ij}}$ for VRPTW in \cite{Cazenave2020VRP}.
In this work, we used $$\beta_{distance}=\frac{-d_{ij}}{max_{kl}(d_{kl})}$$ $$\beta_{lateness}=\frac{-(dd_j - max(d_ij+vt,bt_j))}{biggest\ time\ window}$$ $$\beta{waiting} = 0\ if\ vt+d_{ij}>bt_j$$ $$\beta{waiting} = \frac{-(bt_j - (d_{ij}+vt))}{biggest\ time\ window}\ if\ i\neq depot$$ $$\beta{waiting} = \frac{-(bt_j - max(ftw,d_{ij}+vt))}{biggest\ time\ window}$$ if $i=depot$  

where $d_{ij}$ is the distance between customer $i$ and $j$, $bt_j$ is the beginning of customer $j$ time window, $dd_j$ the end of the time window, $vt$ is the departure instant and $ftw$ is the start of the earliest time window. In the previous formulas, using $-value$ instead of $\frac{1}{value}$ enables zero values for the waiting time or the lateness. To avoid too much influence from $\beta_{waiting}$ at the start of the tour, where the waiting time can be big, we used $max(ftw,d_{ij}+vt)$. The idea is to only take into account the "useful time" lost.
\newpage
\section{Results}
\subsection{Main Results}
In this section, the parameters used for testing NRPA, NRPAD and GNRPA are 3 levels, $\alpha = 1$ and 100 iterations per level. We used NRPA with and without weight initialization as a baseline for GNRPA. We innit the policy with the same formula as in $\beta_{distance}$. For GNRPA the weights were set by a sequential optimization on a subset of the Solomon instances. The weights used are $w_1=15$, $w_2=75$ and $w_3=10$. The results given in table 1 are the best runs out of 10 with different seeds. The running time for NRPA and GNRPA are less than 1800 seconds.

We compare our results with the OR-Tools library. OR-Tools is a Google library for solving optimization problems. It can solve many types of VRP problems, including CVRPTW. Many $first\ solution$ strategies are available. In our experiments, we used "PATH\_CHEAPEST\_ARC" parameter. Starting from a start node, the algorithm connects it to the node which produces the cheapest route segment, and iterates the same process from the last node added. A local search is then performed in order to improve the solution. Several options are also possible here. We used the "GUIDED\_LOCAL\_SEARCH", which uses guided local search to escape local minima. 

Table 1 gives results on the Solomon instances. It compares 4 algorithms to the best known solutions:
\begin{itemize}
\item NRPA : Standard NRPA (3 levels) without policy initialization.
\item NRPAD : Standard NRPA (3 levels) with distance initialization heuristic
\item GNRPA : GNRPA (3 levels) with the bias.
\item OR-Tools : limited to 1800 seconds with parameters described above.
\end{itemize}

The results are compared with the lexicographical approach, first taking into account the number of vehicles used $NV$ and then the total distance traveled $Km$. We observe that NRPA only reached the best score for C101, for which GNRPA and OR-Tools also found the best solution. For all other instances, GNRPA provided better results than NRPA. 

GNRPA performed better than NRPAD on all Solomon instances except R112 thanks to the dynamic bias. 

Although OR-Tools obtains a better result on the majority of the instances (35 out of 56 instances), GNRPA has a better score for 12 of them and an equivalent score for 9 of them. We can see that the results of GNRPA are better on instances of class R1 and RC1 than on instances of class R2 and RC2.  Instances of classes R2 and RC2 have larger time windows.  The time window constraints are thus weakened. When there are fewer constraints, local search performs better than Monte Carlo search. This observation was also made in \cite{Cornu2017LocalSD}. 

 \begin{table*}
    \centering
    \caption{The different algorithms tested on the 56 standard instances}
    \label{my-label}
    \resizebox*{!}{\dimexpr\textheight-2\baselineskip\relax}{%
    \begin{tabular}{|l|lr|lr|lr|lr|lr|}
    \hline
  & NRPA& &NRPAD &  & GNRPA & & OR-Tools & & Best Known&\\
 Instances& NV & Km& NV & Km& NV& Km& NV & Km& NV & Km\\
 \hline
 c101& \textbf{10*} & \textbf{828.94}&\textbf{10*} &\textbf{828.94} & \textbf{10*} & \textbf{828.94}& \textbf{10*} & \textbf{828.94}& 10 & 828.94\\
 \hline
 c102&  10&1,011.40 &10&843.57 &  10& 843.57&  \textbf{10*} & \textbf{828.94} & 10 & 828.94\\
 \hline
 c103&  10&1,105.10 &10&853.82 &  10&843.02 &  \textbf{10*} & \textbf{828.06} & 10 & 828.06\\
 \hline
 c104&  10&1,112.66 &10&857.23 &  \textbf{10}&\textbf{839.96} &  10&846.83 & 10 & 824.78\\
 \hline
 c105&  10&896.93 &\textbf{10*}&\textbf{828.94} &  \textbf{10*}&\textbf{828.94} &  \textbf{10*}&\textbf{828.94} & 10 & 828.94 \\
 \hline
 c106& 10 &853.76 &10&830.54 &  \textbf{10*}&\textbf{828.94} &  \textbf{10*}&\textbf{828.94} & 10 & 828.94\\
 \hline
 c107&  10&891.22 &\textbf{10*}&\textbf{828.94} &  \textbf{10*}&\textbf{828.94} &  \textbf{10*}&\textbf{828.94} & 10 & 828.94\\
 \hline
 c108&  10&1006.69 &10&841.84 &  \textbf{10*}&\textbf{828.94} &  \textbf{10*}&\textbf{828.94} & 10 & 828.94\\
 \hline
 c109&  10&962.35 &10&856.69 &  \textbf{10}&\textbf{834.85} &  10&857.34 & 10 & 828.94\\
 \hline
 c201&  4&709.75 &\textbf{3*}&\textbf{591.56} &  \textbf{3*}&\textbf{591.56} &  \textbf{3*}&\textbf{591.56} & 3 & 591.56\\
 \hline
 c202&  4&929.93 &3&624.04 &  3&611.08 & \textbf{3*}&\textbf{591.56} &  3& 591.56\\
 \hline
 c203&  4&976.00 &3& 620.11&  3&611.79 &  \textbf{3}& \textbf{594.23}&  3& 591.17\\
 \hline
 c204&  4&995.19 &3&619.59 &  3&614.50 &  \textbf{3}& \textbf{593.82}& 3& 590.60\\
 \hline
 c205&  3&702.05 &\textbf{3*}&\textbf{588.88} &  \textbf{3*}&\textbf{588.88} &  \textbf{3*}&\textbf{588.88} &  3& 588.88\\
 \hline
 c206&  4&773.28 &3&591.64 &  \textbf{3*}&\textbf{588.49} &  \textbf{3*}&\textbf{588.49} &  3& 588.49\\
 \hline
 c207&  4&762.73 &3&597.92 &  3&592.50 &  \textbf{3*}&\textbf{588.29} &  3& 588.29 \\
 \hline
 c208&  3&741.98 &3&594.60 &  \textbf{3*}&\textbf{588.32} &  \textbf{3*}&\textbf{588.32} &  3& 588.32\\
 \hline
 r101&  19&1,660.01 &19&1,657.05 &  \textbf{19*}&\textbf{1,650.80} &  19&1,653.15 &  19& 1,650.80\\
 \hline
 r102&  17&1,593.73 &17&1,514.90 &  17&1,508.83 &  \textbf{17}&\textbf{1,489.51} &  17& 1,486.12\\
 \hline
 r103&  14&1,281.89 &14&1,238.34 &  13&1,336.86 &  \textbf{13}&\textbf{1,317.87} &  13& 1,292.68\\
 \hline
 r104&  11&1,098.30 &10&1,029.96 &  10&1,013.62 &  \textbf{10}&\textbf{1,013.23} &  9& 1,007.31\\
 \hline
 r105& 15&1,436.75 &15&1,388.70 &  \textbf{14}&\textbf{1,378.36} &  14&1,393.14 &  14& 1,377.11\\
 \hline
 r106& 12 & 1,364.09 &13 &1,299.46 &  \textbf{12}&\textbf{1,274.47} &  13&1,243.0 &  12& 1,252.03\\
 \hline
 r107& 11 &1,241.15 &11&1,118.32 &  10&1,131.19 &  \textbf{10}&\textbf{1,130.97} &  10& 1,104.66\\
 \hline
 r108& 11 & 1,106.14 &10&992.07 &  10&990.18 &  \textbf{10}& \textbf{963.4}&  9& 960.88\\
 \hline
 r109& 12& 1,271.13 &12&1,185.22 &  12&1,180.09 &  \textbf{12}&\textbf{1,175.48} &  11& 1,194.73\\
 \hline
 r110&  12&1,232.03 &11&1,156.36 &  11&1,140.22 &  \textbf{11}&\textbf{1,125.13} &  10& 1,118.84\\
 \hline
 r111&  12&1,200.37 &11&1,112.51 &  11&1,104.42 &  \textbf{11}&\textbf{1,088.01} &  10& 1,096.72\\
 \hline
 r112&  10&1,162.47 &10&1,005.67 &  10&1,013.50 &  \textbf{10}&\textbf{974.65} &  9&982.14 \\
 \hline
 r201&  5&1,449.95 &5&1,314.92 &  4&1,316.27 &  \textbf{4}& \textbf{1,260.67}&  4& 1,252.37\\
 \hline
 r202&  4&1,335.96 &4&1,173.30 &  4&1,129.89 &  \textbf{4}&\textbf{1,091.66} &  3& 1,191.70\\
 \hline
 r203&  4&1,255.78 &4&1,002.09 &  3&1,004.49 &  \textbf{3}&\textbf{953.85} &  3& 939.50\\
 \hline
 r204&  3&1,074.37 &3&852.66 &  3&787.69 &  \textbf{3}&\textbf{755.01} &  2& 852.52\\
 \hline
 r205&  4&1,299.84 &4&1,067.49 &  3&1,043.81 &  \textbf{3}&\textbf{1,028.6} &  3& 994.43\\
 \hline
 r206&  3&1,270.89 &3&1,129.63 &  3&990.88 &  \textbf{3}& \textbf{923.1}&  3& 906.14\\
 \hline
 r207& 3 &1,215.47 &3&935.98 &  3&900.17 &  \textbf{3}& \textbf{832.82}&  2&890.61 \\
 \hline
 r208& 3 &1,027.12 &3&780.68 &  2&779.25 &  \textbf{2}&\textbf{734.08} &  2&726.82 \\
 \hline
 r209& 4&1,226.67 &4&968.17 &  3&981.82 &  \textbf{3}&\textbf{924.07} &  3& 909.16\\
 \hline
 r210&  4&1,278.61 &4&1,017.99 &  3&995.50 &  \textbf{3}&\textbf{963.4} &  3&939.37 \\
 \hline
 r211&  3&1,068.35 &3&890.02 &  3&850.33 &  \textbf{3}&\textbf{786.28} &  2& 885.71\\
 \hline
 rc101& 15 &1,745.99 &15&1,662.99 &  \textbf{14}&\textbf{1,702.68} &  15&1,639.54 & 14&1,696.95 \\
 \hline
 rc102& 14 & 1,571.50&13&1,525.28 &  \textbf{13}&\textbf{1,509.86} &  13&1,522.89 &  12&1,554.75 \\
 \hline
 rc103&  12&1,400.54 &11&1,313.97 &  \textbf{11}&\textbf{1,287.33} &  12&1,322.84 &  11&1,261.67 \\
 \hline
 rc104&  11&1,264.53 &10&1,178.39 &  10&1,160.55 &  \textbf{10}&\textbf{1,155.33} &  10&1,135.48 \\
 \hline
 rc105&  15&1,620.43 &15&1,554.45 &  \textbf{14}&\textbf{1,587.41} &  14&1,614.98 &  13&1,629.44 \\
 \hline
 rc106&  13&1,486.81 &12&1,413.46 &  \textbf{12}&\textbf{1,397.55} &  13&1,401.73 &  11&1,424.73 \\
 \hline
 rc107& 12 &1,338.18 &11&1,283.18 &  \textbf{11}&\textbf{1,247.80} &  11& 1,255.62&  11&1,230.48 \\
 \hline
 rc108& 11 &1,286.88 &11&1,150.40 &  \textbf{10}&\textbf{1,213.00} &  11&1,148.16 &  10& 1,139.82\\
 \hline
 rc201& 5 &1,638.08 &5&1,407.49 &  4&1,469.50 &  \textbf{4}&\textbf{1,424.01} &  4&1,406.94 \\
 \hline
 rc202& 4 &1,593.54 &5&1,234.92 &  4&1,262.91 &  \textbf{4}& \textbf{1,161.82}&  3& 1,365.65\\
 \hline
 rc203& 4 &1,431.32 &4&1,106.92 &  3&1,123.45 &  \textbf{3}& \textbf{1,095.56}&  3& 1,049.62\\
 \hline
 rc204& 3 &1,260.05 &3&874.79 &  3&864.24 &  \textbf{3}& \textbf{803.06}&  3& 789.46\\
 \hline
 rc205& 5 &1,578.73 &5&1,327.79 &  4&1,347.86 &  \textbf{4}& \textbf{1,315.72}&  4& 1,297.65\\
 \hline
 rc206& 4 &1,412.26 &4&1,175.55 &  3&1,208.52 &  \textbf{3}& \textbf{1,157.2}&  3& 1,146.32\\
 \hline
 rc207& 4 &1,395.02 &4&1,145.46 &  3&1,164.99 &  \textbf{3}&\textbf{1,098.61} &  3& 1,061.14\\
 \hline
 rc208& 3 & 1,182.55&3&970.42 &  3&948.82 &  \textbf{3}& \textbf{843.02}&  3&828.14 \\
 \hline
\end{tabular}%
}
\end{table*}

Figures  \ref{fig:my_label} to  \ref{fig:my_label3} present the evolution of the objective function value over time for 3 algorithms : NRPA, GNRPA and OR-Tools applied to 3 Solomon instances, namely R201, RC101 and RC201. For NRPA and GNRPA the best and the worst runs are represented. First, we notice that the use of GNRPA bias permits a very fast convergence towards a good solution, unlike NRPA without initialization. For example, for instance R201, we can see that within 120 seconds GNRPA yielded better results than NRPA at the end of the run (1500 seconds). We can also notice that the difference between the best and the worst run is much smaller for GNRPA than for NRPA. The convergence speed of GNRPA is slightly worse than the one of OR-Tools. However we see on Figure \ref{fig:my_label2} GNRPA provides a better result for instance RC101 than the result provided by OR-Tools. 

\begin{figure}[H]
    \centering
    \includegraphics[width=0.6\textwidth,keepaspectratio]{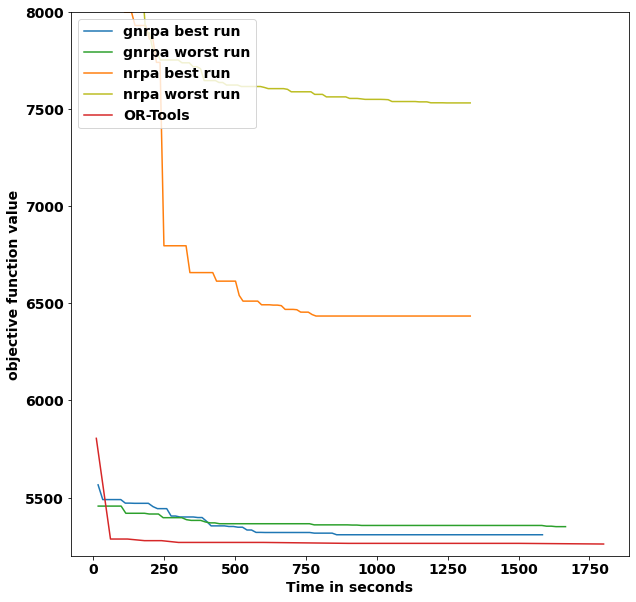}
    \caption{objective function evolution for the different algorithms on instance R201 (objective value as a function of running time)}
    \label{fig:my_label}
\end{figure}

\begin{figure}[H]
    \centering
    \includegraphics[width=0.6\textwidth,keepaspectratio]{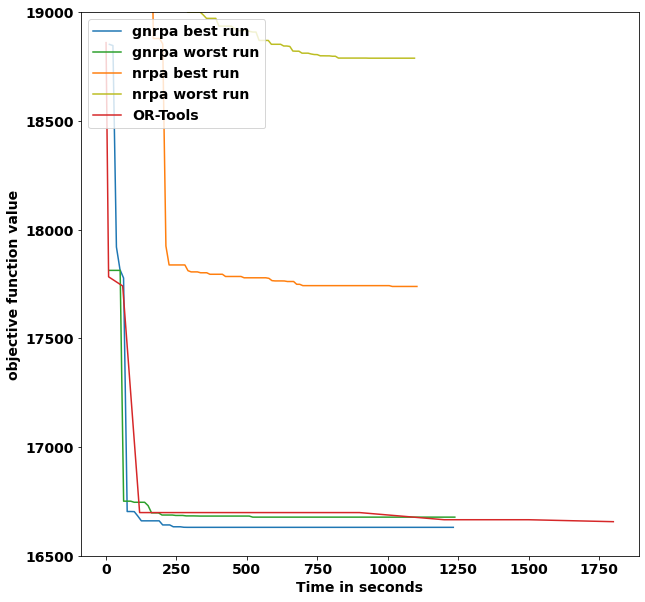}
    \caption{objective function evolution for the different algorithms on instance RC101 (objective value as a function of running time)}
    \label{fig:my_label2}
\end{figure}

\begin{figure}[H]
    \centering
    \includegraphics[width=0.6\textwidth,keepaspectratio]{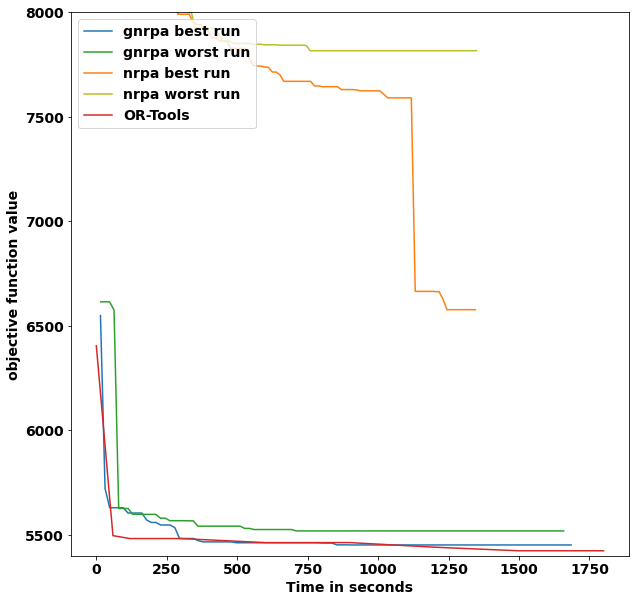}
    \caption{objective function evolution for the different algorithms on instance RC201 (objective value as a function of running time)}
    \label{fig:my_label3}
\end{figure}
\subsection{GNRPA with different number of levels}
In this section, we try to determine whether modifying the number of levels in GNRPA could improve the quality of the solutions. We tried various number of levels $L$ while varying the number of branch per level $N$ so that the total number of playouts at level 0 would remain approximately the
same (1,000,000). Thus, we tested the following values : $L=2$ with $N=1000$, $L=3$ with $N=100$, $L=4$ with $32$, $L=5$ with $N=16$ and $L=6$ with $N=10$. Table 2 presents the results of the different GNRPA parameterizations on the Solomon set of problems. For each instance, 10 runs are performed, and the best one is kept. The results given are the mean value of the objective function over the different instances of the class. It seems that the best setting for GNRPA is $L=3$ or $L=4$. After $L=4$, the results are  worse. This suggests that the trade-off between intensification and diversification is not improved beyond $L=4$.

\begin{table*}
    \centering
    \caption{Mean value of objective functions on Solomon instances for different parameters of GNRPA}
    \label{my-label}
    \begin{tabular}{|l|r|r|r|r|r|r|}
    \hline
  Instances& L=2 N=100,000& L=3 N=100 & L=4 N=32& L=5 N=16 &L=6 N=10 & Best Known\\
 \hline
 C1& 10,851.96 & \textbf{10,838.06}& 10,840.06 & 10,856.55&10,877.58&10,828.22\\
 \hline
 C2& 3,790.65 & 3,605.77& \textbf{3,605.34} & 3,616.54&3,682.11&3,589.85\\
 \hline
 R1& 14,315.26 & 13,838.46& \textbf{13,833.36} & 13,995.11&14,271.23&13,127.00\\
 \hline
 R2& 4,367.52 & \textbf{4,154.73}& 4,173.65 & 4,208.52&4,312.04&3,678,30\\
 \hline
 RC1& 14,239.89 & 13,617.52& \textbf{13,548.72} & 13,670.67&13,945.22&12,884,16\\
 \hline
 RC2&  	5,032.67& 4,667.62& \textbf{4,590.81} & 4,636.56&4,718.91&4,369,24\\
 \hline
\end{tabular}%
\end{table*}

\section{Conclusion}
In this paper, we presented the GNRPA algorithm and its application to the VRP. NRPA and GNRPA have several limitations. First they are less efficient on weakly constrained problems as we presented in section 4.1. Second, NRPA/GNRPA are designed for complete information problems (that is, without random aspects). Finally, the bias must be simple to compute. Indeed, for a GNRPA search it must be calculated $ 100^3 \times \bar{c}$ times, where $\bar{c}$ is the average number of moves considered in the playout function. For the choice of the first travel, 100 moves are taken into account (from the deposit to each of the customers). In order to have a fast and efficient search, the computation of the bias must therefore be simple.\\
The results we obtained show that, provided an appropriate tuning of the GNRPA algorithm, we always outperform the NRPA algorithm on the 56 Solomon instances, while we outperform the NRPAD algorithm on all but one instance (R112). On some instances (12 out of 56) GNRPA also provides better solutions than OR Tools. These preliminary results look promising, so in future work we plan to test some enhancements of the GNRPA such as the Stabilized GNRPA (SGNRPA) : similarly to the Stabilized NRPA, in SGNRPA the Adapt function is not systematically run after each level 1 playout, but with an appropriate periodicity. It also would be interesting to experiment these improvements on larger instances like the "Gehring and Homberger benchmark". Finally, we plan to work on finding better values for the bias weights, or proposing a new bias formula in order to further improve the results. 
%


\bibliographystyle{splncs04}
\bibliography{main}

\end{document}